\documentclass[10pt,conference]{IEEEtran} 
\usepackage{cite}
\usepackage[hyphens]{url}  
\usepackage{balance}   
\usepackage{amsmath,amssymb,amsfonts}

\usepackage{graphicx}
\usepackage{multicol}
\usepackage{float}
\usepackage{textcomp}
\usepackage{booktabs}
\usepackage{tikz}
\usepackage{pgfplots}
\pgfplotsset{compat=1.18}
\usepackage{xcolor}
\usepackage{placeins}
\usepackage{subfig}
\usepackage{algorithm}
\usepackage{algpseudocode}

\usepackage{fancyhdr}
\def\BibTeX{{\rm B\kern-.05em{\sc i\kern-.025em b}\kern-.08em
    T\kern-.1667em\lower.7ex\hbox{E}\kern-.125emX}}

\usepackage{listings}
\usepackage{xcolor}

\lstdefinestyle{instructions}{
    literate={→}{{$\rightarrow$}}1,
    basicstyle=\small\ttfamily,
    backgroundcolor=\color{gray!10},
    frame=single,
    breaklines=true,
    columns=flexible,
    keepspaces=true,
    numbers=none,
    numberstyle=\tiny\color{gray},
    numbersep=5pt,
    showstringspaces=false,
    tabsize=2,
    captionpos=b
}

\begin{document}

\title{Human-in-the-Loop Runbook Improvement with Agentic Support Automation.}
\author{\IEEEauthorblockN{Rocker D'Antonio}
\IEEEauthorblockA{\textit{Amazon Web Services} \\
\textit{Mississippi State University} \\
orcid: 0009-0001-8444-1772}
\and
\IEEEauthorblockN{Harry Xie}
\IEEEauthorblockA{\textit{Amazon Web Services} \\
orcid: 0009-0008-5946-9328}
}

\date{April 2025}

\maketitle 

\begin{abstract}
 Operational support is an important component of production software services. Support requests are often emergent and can come in many forms such as customer escalations or unplanned service interruption. Engineers across organizations have been successful in implementing automation to help streamline support processes but many solutions remain in the hands of human operators. A successful strategy to improve operator response times is to maintain up-to-date runbooks which are documents enumerating triage steps and remediation procedures in the face of identified incidents. This paper is an investigation into an operational system which incorporates Agentic AI into a runbook based incident resolution pattern. The bot uses information in the runbook corpus along with some system-specific tooling to guide the manual operator through the mitigation process then evaluate the outcome and suggest edits to the runbook where it found gaps and in its ability to triage the situation. By using the runbook as a medium to bridge humans and automation, the process maintains explainability and can be decoupled from the agent system. This paper demonstrates a collaboration between human operators and Agentic AI with results from an 17 week study involving 232 production incidents. We conclude with lessons on integrating LLMs into DevOps workflows, related work in AI-assisted operations, and guidance for reproducibility. 
 
\end{abstract}


\section*{Published Version}
This is the accepted version of: R. D'Antonio and H. Xie, "Human-in-the-Loop Runbook Improvement with Agentic Support Automation," \textit{2025 IEEE 7th International Conference on Cognitive Machine Intelligence (CogMI)}, 2025, pp. 341-351, doi: 10.1109/CogMI67134.2025.00046. 

Available: \url{https://ieeexplore.ieee.org/document/11417021}

\section{Introduction}

Operational resilience and timely support are key factors in the success of a software service. Even well-handled incidents can be costly and lead to unpredictable service times, which can erode customer trust.\cite{maxham2002longitudinal} \cite{oconnell_it_outages_2024} 
Engineers are investigating ways to take advantage of new advances in machine learning, generative and agentic artificial intelligence (AI), to improve operational excellence and system resiliency. \cite{zhang2024visionautonomiccomputingllms} \cite{electronics12020296} AI implementation results are mixed, and many organizations are failing to move past the proof of concept stage. \cite{spglobal_ai_2025_ai_poc_fail} The promise of AI seems boundless and, anecdotally, leads to paralyzing scope creep. \cite{bernstein_how_2024} \cite{toczala_successful_2024} In this study, we demonstrate a successful approach to implementing an Agentic AI foundation in operations by augmenting an existing human workflow in a way that humans are encouraged to share knowledge and are not inherently put in competition with AI. The primary strength of this tool is not in its ability to perform tasks that human operators cannot, but rather in its significantly faster execution. This allows human operators to focus on irreplaceable decision-making processes, since humans will inevitably be held accountable to the results.

A typical operations incident starts with an event and then a detection. Next to follow is the response and a short-term remediation, after which a learn phase happens that includes a debrief and proposal for a long-term remediation. \cite{atlassian_incident_management}. In the best cases, the remediation extends to updates to the documentation. \cite{Silva2023DocDebt} While the long-term remediation may fix the current problem, other similar problems may arise that are outside the bands of the fix, leaving the system vulnerable, and the incident response cycle will have to be repeated.  

One well documented approach to resolving operational incidents is with runbooks. \cite{aws_reliability_pillar} \cite{splunk_runbooks} However we cannot expect runbooks to be complete such that every possible variable to an incident can enumerated along with every possible resolution step. We can expect that a human operator has some intuition to fallback on that grants them the ability to make fuzzy judgments and continue to triage to resolution. Large Language Models (LLM) can demonstrate some similar inference. A general description of the LLM inference process is that an embedded input is run through a stack of mathematical calculations called transformers which determine probabilities and eventually distill into an embedded output. \cite{brown2020languagemodelsfewshotlearners} \cite{radford2019language} This probabilistic approach, coupled with an unfathomable matrix of potential probability equations derived from training incomprehensible amounts of data, can yield distilled responses in a vast array of fuzzy situations. Agentic AI leverages the response abilities of LLMs by combining them with tools, often other LLM based agents, and giving them agency to accomplish tasks. \cite{sapkota2025aiagentsvsagentic} 

Incorporating large-language models and agentic AI into incident response is not without its problems. Large language models, as a result of their probabilistic nature, can yield incorrect responses. They are famously known for hallucinating, and are limited in the amount of context they can process at one inference cycle, and may also be unpredictably influenced by the training data that was used to generate the probability equations. \cite{xu2025hallucinationinevitableinnatelimitation} \cite{Heaven2023} Additionally, agentic flow can make irrational decisions based off fuzzy inference leading to indefinite loops, or resource intensive paths that do not yield resolution. These problems can be exacerbated as more tooling options are provided to the agent. \cite{han2024llmmultiagentsystemschallenges} \cite{patil2023gorillalargelanguagemodel} 
While humans can be susceptible to these same issues, they tend to have a better intuition on when to backtrack or stop wasteful processes. \cite{Brooks2021HumanInTheLoop} Research has shown that humans working in the loop with fuzzy systems can lead to better outcomes in many domains.   \cite{WANG2025126971} \cite{tavares2024hitl_survey}

In this study we tackle the following questions: Does embedding an agentic runbook assistant reduce the mean time to resolve (MTTR) incidents? Will a post‑incident gap‑audit agent increases run‑book specificity?

We focused on integrating agentic AI into our previously non-automated incident support workflow. This AI is equipped with access to a Retrieval-Augmented Generation (RAG) knowledge base containing the design documents and runbooks for the application. Our goal was to augment human response ability by following human described procedures with agents and performing previously laborious information gathering. The study had two phases, first we implemented the system to respond to incoming tickets with initial assessment. In the second phase we add the step of asking the system to analyze the gap between it's initial assessment and the documented actions in the ticket that resulted in the resolution. We assume this difference is the presentation of a gap in the knowledge base. From the resulting analysis we tell the agent to suggest edits to the runbook documentation. 

In this paper we detail our implementation, describe the metrics we use for measuring success, provide the results from production study, provide analysis and possible next steps. 

\section{Related Work}
Recent advancements in AI for IT operations (AIOps) suggest promising directions and a number of companies are developing products in this area. Researching publications we found some products trending in a similar direction. For example, PagerDuty’s “Scribe” \cite{pagerduty_scribe} an AI with features that can summarize incident data and, with additional prompting, generate runbook steps. Amazon Q Developer and similar generative AI tools from other companies have shown AI’s ability to draft code and documentation. In addition to product based tools, there are public repositories available with prompts and scripting to help develop run-books, such as AWS's generative-ai-security-runbooks repository on GitHub. \cite{aws_security_runbooks} We did not perform comparison testing directly against these other products. To the best of our knowledge, these systems largely focus on real-time suggestions or initial runbook creation and they do not close the loop after incidents by systematically auditing what was missing and suggesting updates back into the knowledge base. In contrast, our approach introduces a post-incident gap-analysis agent that does exactly this. Our agentic system reviews each incident, identifies runbook deficiencies, and proposes concrete edits which can be quickly converted by a human operator to pull requests.


\section{Methods}

\subsection{Implementation}

The system was implemented so that our operations bot would listen for incoming incident tickets. Incident tickets could be created both manually by customers (customer tickets) and automatically by our monitoring system (alarm tickets). The tickets contain a freeform text description field that explains the incident. For customer tickets, the interface provides a template to the submitter to help structure the description, however this is not enforced. For alarm tickets, the alarming system submits descriptions in a structured way. Details for alarm tickets include timestamps when the threshold has been breached in the monitored period, the breached thresholds, a description of the alarm, and a link to the runbook page which provides triage steps.  

The orchestration agent has access to several tools which it can decide to use while it formulates a response to the incoming ticket. At the foundation of the system is a tool which can retrieve information from the runbook documentation. The runbook has been converted from markdown files to a RAG knowledge base. The second core tool is a memory of previous responses to tickets, in our case implemented in a NoSQL document store. In addition to those core tools are application specific tools which grant access to search various log groups and other components across the application. Since this is a production system, this set of secondary tools was dynamic through out the course of the study. In phase 1 the orchestrator accessed the tools as direct calls to serverless functions by phase 2 these tools had been implemented as agents with their own instructions. 

The orchestrator agent which receives the request is instructed to access the runbook first by using the runbook look up tool. Then it is granted agency to use additional tools before responding back to the ticket with a structured response. (Orchestrator instructions provided in the appendix.) 

In the second phase we added a second orchestration which also listens to the ticket queue but filters messages so it only receives the ticket when it is resolved. This then triggers the agent to do a review. The agent looks back to the initial agent response on the ticket, performs analysis to compare what the actual resolution of the ticket was, and finally comments to the ticket with a gap assessment in the runbook documentation and suggestions for improvement. (Runbook Analysis instructions provided in the appendix.)

\begin{figure}
    \centering
    \includegraphics[width=0.8\linewidth]{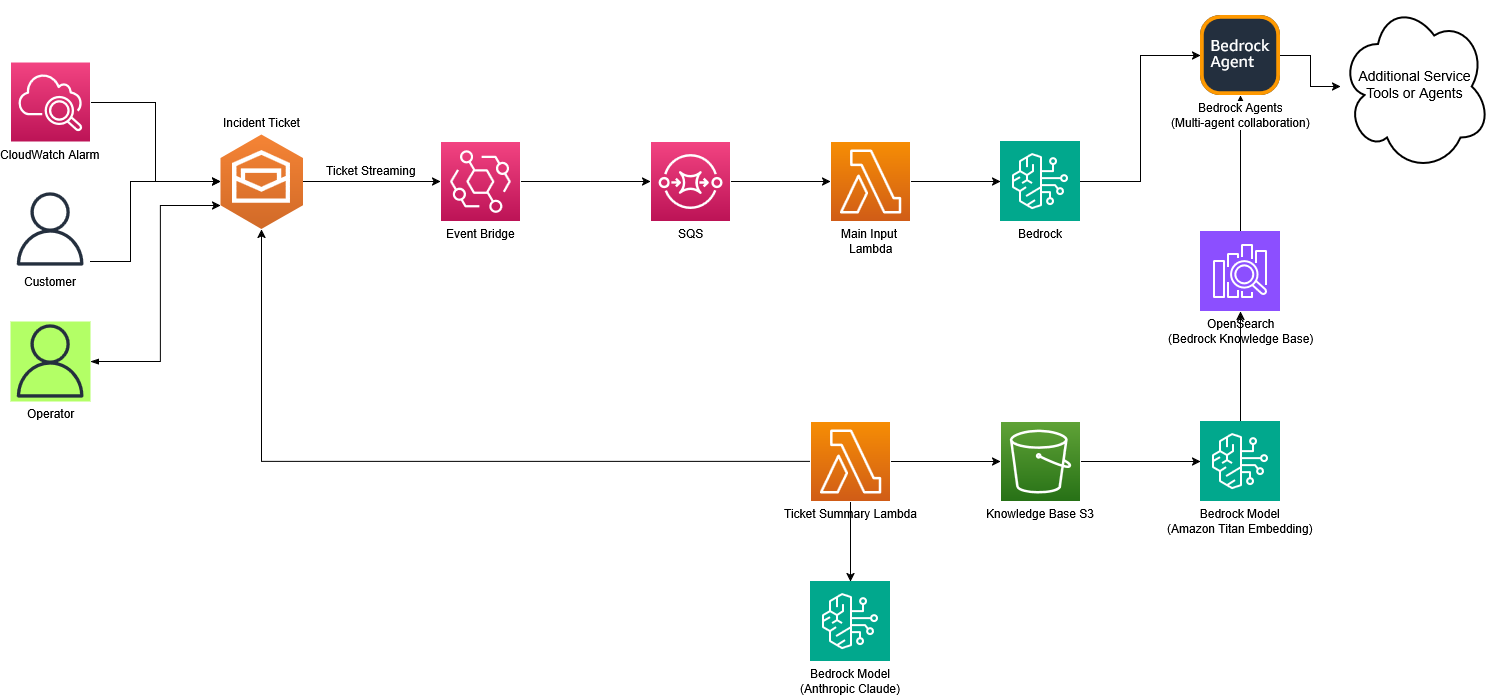}
    \caption{Agent Flow Implementation Details}
    \label{fig:enter-label}
\end{figure}

\subsection{System Overview}
\textbf{Model \& Embeddings}
We used Claude Sonnet 3.7 (200k‑token context, $T=0.2$) for all agent reasoning.  
Document chunks are embedded with AmazonTitan‑E1‑Text‑v2 (1536‑d).  

\textbf{Vector Store}
Our embeddings reside in OpenSearch 2.13 (HNSW: $k=100$, $ef\_search=512$).  
Top‑5 MMR retrieval ($\lambda=0.2$) feeds the orchestrator context.

\textbf{Layered Architecture}
We use a layered architecture.
\begin{itemize}
\item \emph{L0 - Ticket Gateway}: Event Bus \(\rightarrow\) AWS Lambda Function.
\item \emph{L1 - Orchestrator}: Sonnet prompt (see Apendix A) enforces \emph{run‑book‑first}.
\item \emph{L2 - Core tools}: RunBookSearch, TicketMemory.
\item \emph{L3 - Plug‑in agents}: \textsc{LogInspector}, \textsc{MetricQuery}, discovered at runtime.
\item \emph{L4 - Gap‑Audit}: fires on ticket closure, emits suggested runbook edits.
\end{itemize}

\textbf{Operational Footprint}
Median time to first agent comment is 36s (p95 65s); mean incident cost \$0.012.

\subsection{Runbook Organization}
The runbook documentation for this implementation consists of Markdown files contained in a git repository. The runbook is primarily delivered to humans via a rendered static website which is created with MDBook.\cite{mdbook2024} MDBook provides some starting structure including a file called Summary.md in which all runbook pages are indexed as one would index a table of contents for a book. The nested structure of the document allows for files to be chapters and nested sections. When the pages are rendered they flow like a book one unto the next but also can be accessed directly. 
We have structured our documentation into three main sections. First is the design files which detail data models, architecture, and other design choices. The second section contains the runbook procedures. These pages focus on procedures or collections of procedures for maintaining and working with the application along with contacting dependency teams and communicating with customers. The third section contains the alarm triage pages. Every automated alarm must link to a triage page. In the triage page there is a description of the alarm, a high level of how to verify and resolution, then steps to guide the operator, and finally an annotated history of significant related tickets. 
Even with this scaffold, the documentation itself is often organic. Some pages are much larger than others depending on the author and their workload at the time.

\subsection{Description of Study}  

We ran the study in two eight week phases, and involved eight operators. The on-call operators were asked to respond to the requests from the agent for updating the runbook in a timely manner. Updating documentation based on incident findings is an expected but not enforced part of the on-call work allocation, and has a low priority relative to mitigating active issues. In our study the on-call operators were selected from the development team for the software service, which is the typical policy at the company where the study was implemented. We note this because the operators may or may not have additional knowledge of the software service learned from developing it directly and attempts to replicate the study should take that into account.

\subsection{Deployment Setting}

The deployment was conducted with the software engineering team responsible for an online data reporting service. The team had eight on-call engineers rotating weekly. Prior to introducing the AI, the team followed standard incident response practices: using a paging system, Slack for coordination, and referencing an internal wiki for runbooks. We injected our agentic operations AI for eight weeks of treatment, during which the AI agent was active on all incidents. All engineers received a brief training on how to read the agent output and how to review its documentation PRs. We emphasized that the AI was a copiloting tool and that they remained the decision-makers (addressing any potential reluctance). Throughout, the same set of services and incident types were in scope, to ensure comparability.

\subsection{Measurements}

The study was designed to determine the efficacy of a human in the loop runbook update process driven by agentic AI operations. The system was implemented in a live production environment. In order to determine the efficacy, we set up several measurements to account for the unpredictable nature of incident response in production which makes it difficult to assess empirically. 

\subsubsection{Operational Metrics} 

The first metric that we tracked was the rate of runbook update code reviews (CR) in relationship to incoming tickets. From this metric we can determine whether having the agent suggest runbook updates would increase the number of updates made by the operators. Since we are monitoring a production system it is possible this metric can be influenced by other adjacent development work unrelated to our experiment, however we make the assumption those outside influences remain consistent enough so that we can still gain insights.

The second metric, mean time to resolve (MTTR) was selected to examine the change in the resolution time of issues over the course of the study. MTTR is a common key performance indicator (KPI) for studying support processes. \cite{atlassian_incident_metrics} \cite{atlassian_incident_management} \cite{splunk_incident_metrics} We do, however, recognize we would need some significant length of time in an event-driven, uncontrolled study to get a real determination of the impact.

\subsubsection{Textual Metrics} 

In order to have relevant metrics to evaluate the runbook text we first started with a voice of the customer approach, collecting pain points related to runbook documentation from the operators. The three highest pain-points were: not enough documentation (i.e. Volume), disorganized documentation made information hard to find (i.e. Order) and the documentation was too vague (i.e. Specificity). Since the documentation was stored in a repository, we were able to statically analyze snapshots of text as we entered and concluded the study phases.

We decided to use word count as a measure of Volume, header count as measure of Order, and a rudimentary average Term Frequency - Inverse Document Frequency (TF-IDF) score for Specificity. These were all easily obtained without complex implementation. Header count refers to the sectional headers in the documents. In our case, since our documentation corpus was composed in Markdown, we counting header syntax matching the regular expression: \texttt{'\#+ '}. 

Specificity and vagueness can be measured in documents in numerous ways which are worthy of study on their own accord. \cite{uday_vagueness} \cite{icard2023measuringvaguenesssubjectivitytexts} \cite{debnath-roth-2021-computational} We use TF-IDF, with max normalization and smoothing, to calculate the importance of terms, minus stop words, across the document and then generate an average score across the document. \cite{sparck1972statistical} \cite{aizawa2003information} \cite{beals2022analyzing} This gives us some general idea of specificity by allowing us to track the importance of the terms to documents across the corpus. We interpret that higher TD-IDF average score should indicate that individual runbook files are containing more specific instructions. The algorithm filters 134 stop words and applies tokenization using regex pattern \texttt{\b[a-z0-9-.:/]+\b}.

The TF-IDF average algorithm works such that given a corpus $D = \{d_1, d_2, ..., d_N\}$ of $N$ documents and a vocabulary $V$ of unique terms,
the TF-IDF score for term $t$ in document $d$ is calculated as:

\begin{equation}
\text{TF-IDF}(t,d) = \text{TF}(t,d) \times \text{IDF}(t)
\end{equation}

Where:

\begin{enumerate}
\item Term Frequency (TF) with max normalization:
    \begin{equation}
    \text{TF}(t,d) = \frac{f_{t,d}}{\max_{t' \in d} f_{t',d}}
    \end{equation}
    where $f_{t,d}$ is the raw frequency of term $t$ in document $d$.

\item Inverse Document Frequency (IDF) with smoothing:
    \begin{equation}
    \text{IDF}(t) = 1 + \log\left(1 + \frac{N + 1}{\text{df}(t) + 1}\right)
    \end{equation}
    where $\text{df}(t)$ is the document frequency (number of documents containing term $t$).

\item Corpus-level TF-IDF aggregation:
    \begin{equation}
    \text{TF-IDF}_{\text{corpus}}(t) = \sum_{d \in D} \text{TF-IDF}(t,d)
    \end{equation}

\item Document-level average score:
    \begin{equation}
    \text{Score}(d) = \frac{1}{|V_d|} \sum_{t \in V_d} \text{TF-IDF}(t,d)
    \end{equation}
    where $V_d$ is the vocabulary of document $d$.
\end{enumerate}

\section{Results}

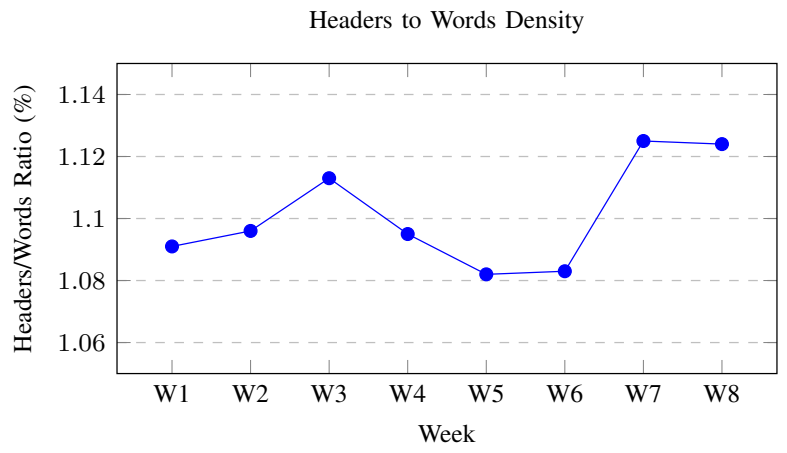
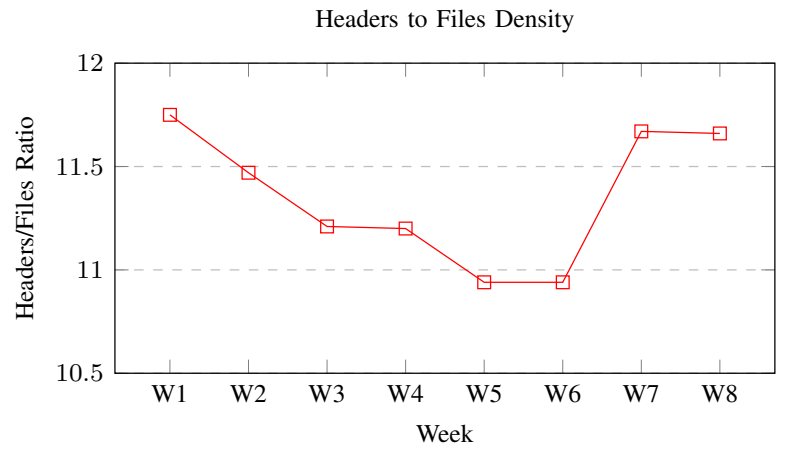
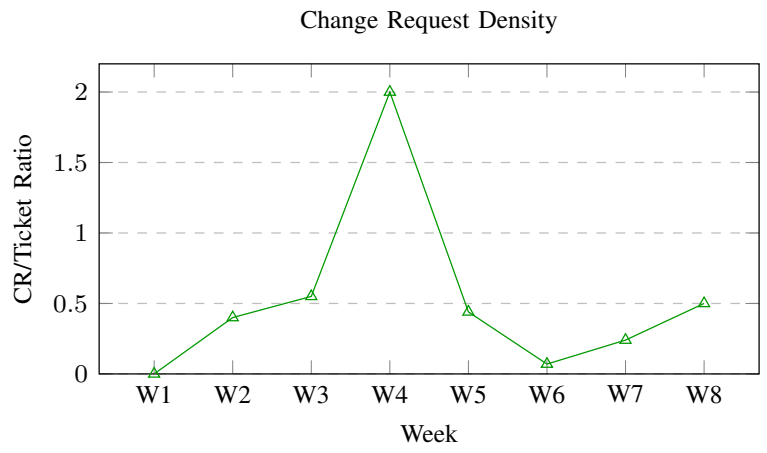
\begin{figure}[H]
\centering
\subfloat[Headers to Words Ratio Over Time]{
\begin{tikzpicture}[scale=0.6]
\begin{axis}[
    xlabel={Week},
    ylabel={Headers/Words Ratio (\%)},
    title={Headers to Words Density},
    ymajorgrids=true,
    grid style=dashed,
    width=\columnwidth,
    height=5cm,
    xtick={1,2,3,4,5,6,7,8},
    xticklabels={W1,W2,W3,W4,W5,W6,W7,W8},
    tick label style={font=\footnotesize},
    label style={font=\footnotesize},
    title style={font=\footnotesize},
    ymin=1.05,
    ymax=1.15
]

\addplot[
    color=blue,
    mark=*,
    ]
    coordinates {
    (1,1.091)
    (2,1.096)
    (3,1.113)
    (4,1.095)
    (5,1.082)
    (6,1.083)
    (7,1.125)
    (8,1.124)
    };

\end{axis}
\end{tikzpicture}
\label{fig:words-density}
}

\vspace{2mm}

\subfloat[Headers to Files Ratio Over Time]{
\begin{tikzpicture}[scale=0.6]
\begin{axis}[
    xlabel={Week},
    ylabel={Headers/Files Ratio},
    title={Headers to Files Density},
    ymajorgrids=true,
    grid style=dashed,
    width=\columnwidth,
    height=5cm,
    xtick={1,2,3,4,5,6,7,8},
    xticklabels={W1,W2,W3,W4,W5,W6,W7,W8},
    tick label style={font=\footnotesize},
    label style={font=\footnotesize},
    title style={font=\footnotesize},
    ymin=10.5,
    ymax=12
]

\addplot[
    color=red,
    mark=square,
    ]
    coordinates {
    (1,11.75)
    (2,11.47)
    (3,11.21)
    (4,11.20)
    (5,10.94)
    (6,10.94)
    (7,11.67)
    (8,11.66)
    };

\end{axis}
\end{tikzpicture}
\label{fig:files-density}
}

\vspace{2mm}

\subfloat[Change Request to Ticket Ratio Over Time]{
\begin{tikzpicture}[scale=0.6]
\begin{axis}[
    xlabel={Week},
    ylabel={CR/Ticket Ratio},
    title={Change Request Density},
    ymajorgrids=true,
    grid style=dashed,
    width=\columnwidth,
    height=5cm,
    xtick={1,2,3,4,5,6,7,8},
    xticklabels={W1,W2,W3,W4,W5,W6,W7,W8},
    tick label style={font=\footnotesize},
    label style={font=\footnotesize},
    title style={font=\footnotesize},
    ymin=0,
    ymax=2.2
]

\addplot[
    color=green!60!black,
    mark=triangle,
    ]
    coordinates {
    (1,0.00)
    (2,0.40)
    (3,0.55)
    (4,2.00)
    (5,0.44)
    (6,0.07)
    (7,0.24)
    (8,0.50)
    };

\end{axis}
\end{tikzpicture}
\label{fig:cr-ratio}
}
\caption{P1 Documentation and Change Request Metrics}
\label{fig:all-metrics}
\end{figure}

In regard to operational metrics, over the eight week Phase 1 study we had nine customer tickets and 68 alarm tickets. We saw the MTTR for both ticket types trend positively downward after Week 3 as shown in Table \ref{tab:tickets}. This could indicate efficacy of the implemented system. However, though runbook updates were notched starting in Week 2 there is no clear indicator using the CR/Ticket metric (Fig. \ref{fig:cr-ratio}) that the update frequency is driven by ticket count. In Phase 2, which had 135 alarm tickets and 20 customer tickets over nine weeks, the trend continues where the CR/Ticket metric remains less than one to one. 

One consideration when reviewing this data is that, particularly for auto-cut alarms, one incident could create more than one ticket which could skew the ratio. We left the data in the raw form as we were not confident that we could accurately identify duplications, and the automation system itself acts on every ticket. A second consideration is that MTTR involves many external factors. In retrospect, we would have liked to have implemented a Mean Time To Diagnosis metric, however the ticketing system was not implemented to track this metric at the time of our study. 

As for our textual metrics, the volume of documentation steadily increased through out the study but we didn't see a noticeable upward trend in header to word (Fig. \ref{fig:words-density}) or header to files (Fig. \ref{fig:files-density}) ratios to indicate an increase in order in the documentation. We saw our TF-IDF average decreased slightly in Phase 1, as shown in Table \ref{tab:doc-metrics} and again in Phase 2, as shown in Table \ref{tab:doc-metrics2}.  

While there is a small trend in TF-IDF average across the phases, we are reluctant to read too much into the difference and correlate it as an indicator of the vagueness of the documentation. TF-IDF gives high scores to terms that are used frequently in few documents, and very low scores for terms that are used infrequently in few documents. For example, in Phase 1 Week 8, the term "data", which appears in a few documents and many times in the data dictionary document, had a TD-IDF score of 50.0917. In that same week, a column name of one table, which appears one time only and only in the data dictionary document, has a TD-IDF score of .0082. So there is a question of how to interpret these values for runbook efficacy prediction, a specific term in a single document indicates the documentation is specific. Terms having high scores, read as having high importance to few documents, can indicate good organization of the documentation.    

Finally, in the operator retrospective at the end of the first phase there was anecdotal praise for the effectiveness of the implementation in helping operators resolve issues in a more timely manner. In the retrospective after the second phase there was a push for long term adoption of the system by the team and interest from other teams.

\subsection{Phase 1 Tables and Charts}

\begin{table}[H]
\centering
\begin{tabular}{lrrrrrr}
\toprule
 & \multicolumn{3}{c}{Alarm Tickets} & \multicolumn{3}{c}{Customer Tickets} \\
\cmidrule(lr){2-4} \cmidrule(lr){5-7}
Week & Count & Open* & MTTR* & Count & Open* & MTTR* \\
\midrule
Week1 & 7 & 19 & 2.71 & 0 & 0 & 0.00 \\
Week2 & 5 & 29 & 5.80 & 0 & 0 & 0.00 \\
Week3 & 11 & 142 & 12.91 & 0 & 0 & 0.00 \\
Week4 & 2 & 13 & 6.50 & 1 & 14 & 14.00 \\
Week5 & 6 & 22 & 3.67 & 3 & 28 & 9.33 \\
Week6 & 14 & 22 & 1.57 & 1 & 7 & 7.00 \\
Week7 & 14 & 33 & 2.36 & 3 & 16 & 5.33 \\
Week8 & 9 & 6 & 0.67 & 1 & 5 & 5.00 \\
\bottomrule
\end{tabular}
\caption{P1 Weekly Ticket Metrics. *Days rounded down.}
\label{tab:tickets}
\end{table}

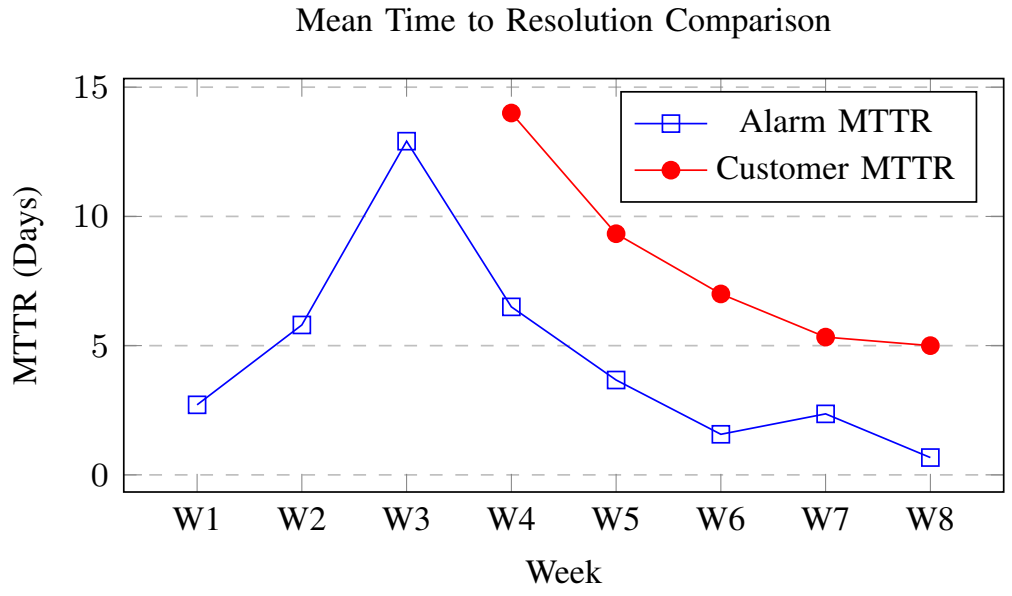
\begin{figure}[H]
\centering
\begin{tikzpicture}[scale=0.8]  
\begin{axis}[
    xlabel={Week},
    ylabel={MTTR (Days)},
    title={Mean Time to Resolution Comparison},
    legend pos=north east,
    ymajorgrids=true,
    grid style=dashed,
    width=\columnwidth,    
    height=5cm,           
    xtick={1,2,3,4,5,6,7,8},
    xticklabels={W1,W2,W3,W4,W5,W6,W7,W8},  
    tick label style={font=\footnotesize},    
    label style={font=\footnotesize},         
    title style={font=\footnotesize},         
    legend style={font=\footnotesize}         
]

\addplot[
    color=blue,
    mark=square,
    ]
    coordinates {
    (1,2.71)
    (2,5.80)
    (3,12.91)
    (4,6.50)
    (5,3.67)
    (6,1.57)
    (7,2.36)
    (8,0.67)
    };
    
\addplot[
    color=red,
    mark=*,
    ]
    coordinates {
    (4,14)
    (5,9.33)
    (6,7)
    (7,5.33)
    (8,5)
    };

\legend{Alarm MTTR,Customer MTTR}  
\end{axis}
\end{tikzpicture}
\caption{P1 MTTR between Alarm and Customer Tickets}
\label{fig:mttr-comparison}
\end{figure}

\begin{table}[H]
\centering
\begin{tabular}{lrrrrrr}
\toprule
Week & Files & Words & Headers & TF-IDF & CRs & CR/Tickets \\
\midrule
Week1 & 57 & 61,392 & 670 & .5792 & 0 & 0.00 \\
Week2 & 59 & 61,782 & 677 & .5958 & 2 & 0.40 \\
Week3 & 62 & 62,437 & 695 & .6069 & 6 & 0.55 \\
Week4 & 65 & 66,491 & 728 & .5986 & 6 & 2.00 \\
Week5 & 68 & 68,746 & 744 & .5987 & 4 & 0.44 \\
Week6 & 68 & 68,731 & 744 & .5986 & 1 & 0.07 \\
Week7 & 72 & 74,654 & 840 & .5783 & 4 & 0.24 \\
Week8 & 73 & 75,662 & 851 & .5756 & 5 & 0.50 \\
\bottomrule
\end{tabular}
\caption{P1 Documentation and Change Request Metrics}
\label{tab:doc-metrics}
\end{table}

\subsection{Phase 2 Tables and Charts}

\begin{table}[H]
\centering
\begin{tabular}{lrrrrrr}
\toprule
 & \multicolumn{3}{c}{Alarm Tickets} & \multicolumn{3}{c}{Customer Tickets} \\
\cmidrule(lr){2-4} \cmidrule(lr){5-7}
Week & Count & Open* & MTTR* & Count & Open* & MTTR* \\
\midrule
Week1 & 18 & 85 & 4.72 & 5 & 50 & 10.00 \\
Week2 & 9 & 27 & 3.00 & 2 & 6 & 3.00 \\
Week3 & 11 & 58 & 5.27 & 2 & 4 & 2.00 \\
Week4 & 20 & 47 & 2.35 & 5 & 54 & 10.80 \\
Week5 & 11 & 48 & 4.36 & 2 & 31 & 15.50 \\
Week6 & 20 & 19 & 0.95 & 1 & 18 & 18.00 \\
Week7 & 26 & 229 & 8.81 & 1 & 21 & 21.00 \\
Week8 & 11 & 88 & 8.00 & 1 & 6 & 6.00 \\
Week9 & 9 & 30 & 3.33 & 1 & 1 & 1.00 \\
\bottomrule
\end{tabular}
\caption{P2 Weekly Ticket Metrics. *Days rounded down.}
\label{tab:tickets2}
\end{table}

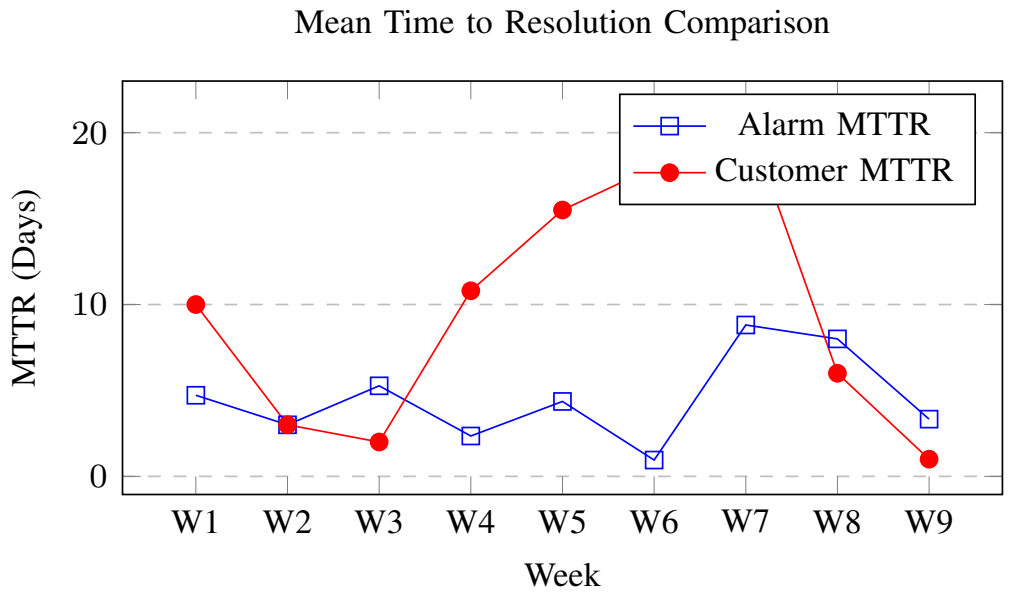
\begin{figure}[H]
\centering
\begin{tikzpicture}[scale=0.8]  
\begin{axis}[
    xlabel={Week},
    ylabel={MTTR (Days)},
    title={Mean Time to Resolution Comparison},
    legend pos=north east,
    ymajorgrids=true,
    grid style=dashed,
    width=\columnwidth,    
    height=5cm,           
    xtick={1,2,3,4,5,6,7,8,9},
    xticklabels={W1,W2,W3,W4,W5,W6,W7,W8,W9},  
    tick label style={font=\footnotesize},    
    label style={font=\footnotesize},         
    title style={font=\footnotesize},         
    legend style={font=\footnotesize}         
]

\addplot[
    color=blue,
    mark=square,
    ]
    coordinates {
    (1,4.72)
    (2,3.00)
    (3,5.27)
    (4,2.35)
    (5,4.36)
    (6,0.95)
    (7,8.81)
    (8,8.00)
    (9,3.33)
    };
    
\addplot[
    color=red,
    mark=*,
    ]
    coordinates {
    (1,10.00)
    (2,3.00)
    (3,2.00)
    (4,10.80)
    (5,15.50)
    (6,18.00)
    (7,21.00)
    (8,6.00)
    (9,1.00)
    };

\legend{Alarm MTTR,Customer MTTR}  
\end{axis}
\end{tikzpicture}
\caption{P2 MTTR between Alarm and Customer Tickets}
\label{fig:mttr-comparison2}
\end{figure}

\begin{table}[H]
\centering
\begin{tabular}{lrrrrrr}
\toprule
Week & Files & Words & Headers & TF-IDF & CRs & CR/Tickets \\
\midrule
Week1 & 75 & 75,892 & 857 & .5764 & 5 & 0.22 \\
Week2 & 75 & 75,892 & 857 & .5764 & 0 & 0.00 \\
Week3 & 76 & 79,223 & 891 & .5697 & 5 & 0.38 \\
Week4 & 78 & 80,898 & 904 & .5647 & 3 & 0.12 \\
Week5 & 80 & 81,403 & 917 & .5704 & 3 & 0.23 \\
Week6 & 80 & 81,403 & 917 & .5704 & 0 & 0.00 \\
Week7 & 80 & 81,652 & 918 & .5693 & 2 & 0.07 \\
Week8 & 80 & 81,833 & 921 & .5687 & 1 & 0.08 \\
Week9 & 80 & 81,858 & 921 & .5687 & 2 & 0.20 \\
\bottomrule
\end{tabular}
\caption{P2 Documentation and Change Request Metrics}
\label{tab:doc-metrics2}
\end{table}

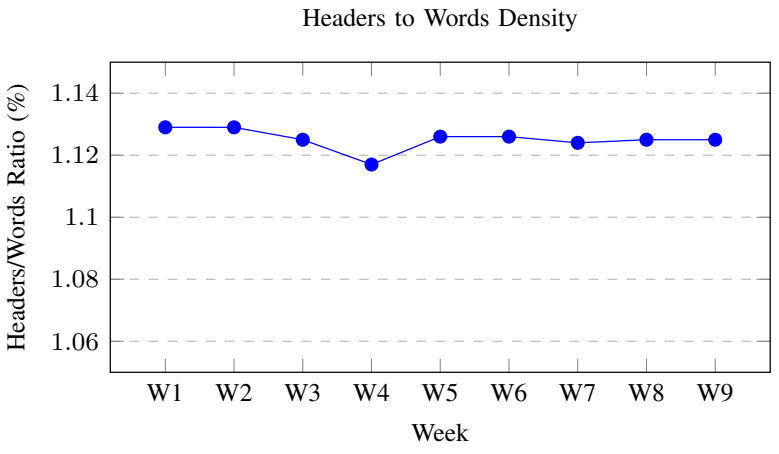
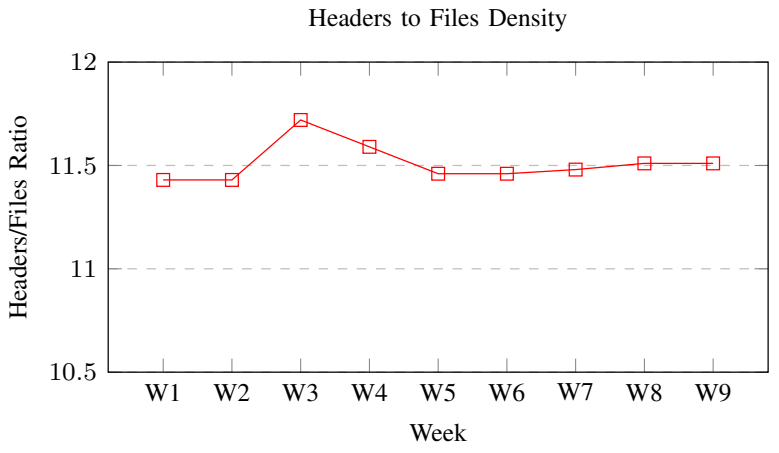
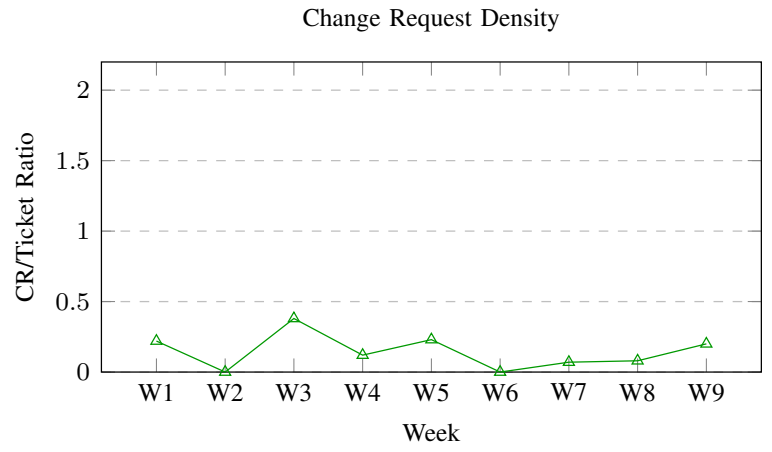
\begin{figure}[H]
\centering
\subfloat[Headers to Words Ratio Over Time]{
\begin{tikzpicture}[scale=0.6]
\begin{axis}[
    xlabel={Week},
    ylabel={Headers/Words Ratio (\%)},
    title={Headers to Words Density},
    ymajorgrids=true,
    grid style=dashed,
    width=\columnwidth,
    height=5cm,
    xtick={1,2,3,4,5,6,7,8,9},
    xticklabels={W1,W2,W3,W4,W5,W6,W7,W8,W9},
    tick label style={font=\footnotesize},
    label style={font=\footnotesize},
    title style={font=\footnotesize},
    ymin=1.05,
    ymax=1.15
]

\addplot[
    color=blue,
    mark=*,
    ]
    coordinates {
    (1,1.129)
    (2,1.129)
    (3,1.125)
    (4,1.117)
    (5,1.126)
    (6,1.126)
    (7,1.124)
    (8,1.125)
    (9,1.125)
    };

\end{axis}
\end{tikzpicture}
\label{fig:words-density2}
}

\vspace{2mm}

\subfloat[Headers to Files Ratio Over Time]{
\begin{tikzpicture}[scale=0.6]
\begin{axis}[
    xlabel={Week},
    ylabel={Headers/Files Ratio},
    title={Headers to Files Density},
    ymajorgrids=true,
    grid style=dashed,
    width=\columnwidth,
    height=5cm,
    xtick={1,2,3,4,5,6,7,8,9},
    xticklabels={W1,W2,W3,W4,W5,W6,W7,W8,W9},
    tick label style={font=\footnotesize},
    label style={font=\footnotesize},
    title style={font=\footnotesize},
    ymin=10.5,
    ymax=12
]

\addplot[
    color=red,
    mark=square,
    ]
    coordinates {
    (1,11.43)
    (2,11.43)
    (3,11.72)
    (4,11.59)
    (5,11.46)
    (6,11.46)
    (7,11.48)
    (8,11.51)
    (9,11.51)
    };

\end{axis}
\end{tikzpicture}
\label{fig:files-density2}
}

\vspace{2mm}

\subfloat[Change Request to Ticket Ratio Over Time]{
\begin{tikzpicture}[scale=0.6]
\begin{axis}[
    xlabel={Week},
    ylabel={CR/Ticket Ratio},
    title={Change Request Density},
    ymajorgrids=true,
    grid style=dashed,
    width=\columnwidth,
    height=5cm,
    xtick={1,2,3,4,5,6,7,8,9},
    xticklabels={W1,W2,W3,W4,W5,W6,W7,W8,W9},
    tick label style={font=\footnotesize},
    label style={font=\footnotesize},
    title style={font=\footnotesize},
    ymin=0,
    ymax=2.2
]

\addplot[
    color=green!60!black,
    mark=triangle,
    ]
    coordinates {
    (1,0.22)
    (2,0.00)
    (3,0.38)
    (4,0.12)
    (5,0.23)
    (6,0.00)
    (7,0.07)
    (8,0.08)
    (9,0.20)
    };

\end{axis}
\end{tikzpicture}
\label{fig:cr-ratio2}
}
\caption{P2 Documentation and Change Request Metrics}
\label{fig:all-metrics2}
\end{figure}

\section{Discussion}

Our feelings about the study and it's results are optimistic. While the metrics we tracked for our short study didn't provide inescapably convincing results to prove that the human in the loop runbook improvement with agentic support automation system compelled operators to make more frequent and higher quality runbook updates, we observed a positive downward trend in MTTR and received anecdotal feedback from the operators that the system is an improvement from their previous workflow. We also didn't see any obvious regressions in operational or textual statistics. We have a proven initial deployment scope which will benefit others interested in implementing similar systems. There are many immediate actions that we can take to expand the system and it is modular enough that future technological advances can be integrated without great expense.

\subsection{Lessons Learned}

After several iterations we have acquired preferences on how to implement the key components. We suggest that the incident should be processed through a ticketing system that has programatically searchable history. These tickets can be either triggered by some sort of monitoring system or created manually as support requests. There should be a direct link from the ticketing system to the agent so that the agent can be initiated by the incoming ticket. We further suggest the agent has a link to the ticket so it can continue to comment and respond to comments, this can serve as a place for it to store state and to interact with the operator. 
 
The runbook documentation should be treated as code and exist in one or more repositories with a code review process. This allows for proper revision history, CI/CD on the knowledge base, and will provide the ability for the agent to make suggestions for edits directly that can then be reviewed in line with the existing software process. It is important to think about the runbook corpus as a source of decision making logic for the agent, and should be treated the same way one might treat code that handles business logic.

Restrict the agent from accessing it's own feedback directly. If it is able to produce wrong feedback and read it then it will convince itself to suggest bad solutions. 

The agentic system can be more robust and include additional system-specific tooling. We suggest including the ability to query the status of the software service and its dependencies, the ability to review past tickets, and the ability to inspect relevant logs. However, in order to keep with the human operator paradigm, all of these abilities must also be accessible to the operator and describable in the runbook.

\subsection*{Strengths and Benefits}

The primary benefit observed was operational speed and efficiency without sacrificing safety. Incidents were resolved faster, as evidenced by the reduction in MTTR, and this was achieved by accelerating human work such as suggesting known fixes, automating checks rather than relying on opaque ``black-box'' automation. Th operators still followed procedure; the AI simply made following procedure easier.

Another strength was the system’s support for continuous improvement of the knowledge base. Traditionally, process improvement in operations is retrospective and often sporadic. Here, we had a near real-time loop where every incident immediately produced a knowledge artifact. This promotes a culture of constant learning.

Interestingly, even in cases where the AI found no gap (i.e., the runbook was already complete), that in itself served as a useful signal in that it reinforced operator confidence that the documentation was solid and comprehensive.

The system also led to greater consistency of response. Different engineers, when assisted by the same runbook suggestions, tended to converge on similar resolution approaches. This reduction in outcome variance improves reliability and predictability in incident response.

Another notable benefit was the low effort required to maintain documentation with AI assistance. Team leads noted that updating runbooks used to be a deprioritized chore. With the AI, relevant changes were suggested proactively and easily converted to pull requests, dramatically lowering the barrier to documentation upkeep. As runbooks became more complete over time, the effort required per incident decreased creating a compounding benefit.

Lastly, the system aided in onboarding and training New team members who learned procedures more quickly by seeing the AI’s step-by-step suggestions during live incidents, and reading the updated documentation afterward. This aligns with the goals of tools like Nissist \cite{an2024nissistincidentmitigationcopilot}, which aim to reduce onboarding fatigue by converting unstructured guidance into structured, actionable knowledge.

\subsection{Limitations}

Despite the overall success, there are important limitations to acknowledge:

\begin{itemize}
    \item Scope of Knowledge: The system is only as good as the knowledge in the runbooks and what it can infer from incidents. In a few incidents, the root cause was entirely novel (e.g., a new type of failure no one had seen). In such cases, the AI’s contributions were minimal besides generic advice, and it sometimes gave an irrelevant suggestion from a loosely related runbook. For example, an incident caused by a rogue internal library had no documentation; the AI retrieved a tangential “dependency issues” runbook which wasn’t very helpful. The on-call solved it via intuition and afterward wrote a new runbook (the AI couldn’t because it had no reference). Over time, this will happen less as more scenarios are covered, but the system is not omniscient and it cannot magically handle completely unknown problems.

    \item Quality of AI Suggestions: While most suggestions were accepted, not all were perfect. One example: the AI recommended adding a step to check a log file that turned out not to exist in that context (it over-generalized from another service’s runbook). This suggestion was caught in review and discarded. This highlights that the AI can still hallucinate or make incorrect assumptions if the retrieved context is slightly off. We mitigated this by including incident-specific information in the prompt, but it’s not foolproof. Fortunately, the human review step prevented any incorrect info from making it into the documentation. In practice, this is manageable, but it means the AI is not at a stage where it can update knowledge fully autonomously with 100\% accuracy.

    \item CR Overload: With CR to ticket ratio being tracked, one might fear spam of CRs. In our 17 week study, 28 CRs were created. This is a manageable review load for an eight person engineering team. If the system were at scale across many teams, or the volume of incoming tickets were much higher the CR load may become less manageable. There is a risk that if the volume of CRs is too high, humans start ignoring them. Our recommendation is to consolidate changes when possible (the agent could batch minor fixes into a single weekly CR per runbook, for example). We intentionally limited the scope to one peer review per incident to not complicate things but in future a smarter batching might be needed.

    \item Long-Term Maintenance: We have not yet tested the long-term (months, years) impact. Over time, runbooks may accumulate a lot of AI suggested content. There is a possibility of bloat or that some suggestions become outdated as systems change. Ideally, the same agent could also identify obsolete info (a form of cleanup), but we did not implement that. So there is a maintenance consideration: having the human in the loop to review will help with bloat but another mechanism may need to be implemented to address stale advice. Some participants suggested that we schedule the agent to periodically audit runbooks against current system behavior.

    \item Management Policy: AI suggestions from the agentic system were a positive force for addressing runbook gaps however they can not replace a policy mandating or rewarding runbook updates. We saw the weeks with the highest amount of tickets have lower CR/Ticket ratios like Phase 2 Weeks 6 \& 7 (see \ref{tab:doc-metrics2}). This reduction matches intuition; more work less time to document. On the other hand, we saw weeks with low tickets and low runbook update CRs like Phase 2 Week 2 \& 8 (see \ref{tab:doc-metrics2}.) 
    
\end{itemize}

\section{Conclusion}

There are a number of different visions for how AI can be used in the operations support scenario, the most ambitious of which is complete automation of operations workflows by AI agents. The state of the industry suggests that building systems with this expectation is prone to failure. We have demonstrated that a useful system can be built by examining the human-based workflow of a support operator and augmenting that flow with Agentic AI. By creating a cooperative relationship between the agentic system and the human, the system is able to ask the for the additional knowledge it needs to succeed and the human operator is able to direct that learning. 

Future work in this area may focus on enabling the agentic system to be more specific in it's prompting for knowledge and suggestions to the runbook corpus. This could include directly submitting CRs to the documentation code base from the context of the incident. Another area of interest is how to implement knowledge exchange between agentic systems so that domain expertise can be containerized into the smaller implemented systems but still called upon by the agents of of dependencies.  

At the core of the human in the loop agentic AI system is an exchange of feedback between an artificial intelligence and a human operator. The human is ultimately accountable for the results of the system. They artificial intelligence is able to recognize gaps in the knowledge base but the human will be the teacher. Since the process is modeled after human activity judgment can be applied in a natural way to what gets taught. We have demonstrated in our study that a system can be implemented beneficially in production environments and expanded iteratively.

\section*{Acknowledgment}

We would like to thank and acknowledge Sasikala Singamaneni who is always pushing us to raise the bar and removing obstacles in our path. Rocker would also like to thank Dr. Zhiqian Chen for teaching the difference between experimenting and being an academic. We would like to thank our friends, family, and coworkers who supported us while we wrote the paper.

\balance
\bibliographystyle{IEEEtran}
\bibliography{IEEEabrv,reference}

\section*{Appendix A}
\subsection*{Orchestrator Prompt}
\begin{lstlisting}[style=instructions]
<instructions>
You are a supervisor agent coordinating a multi-agent workflow for troubleshooting on-call issues. Your responses must be in clear markdown format suitable for ticket comments. You handle system alerts and customer inquiries, delegating to specialized agents and synthesizing their findings into actionable insights.

## Ticket Assessment Process

When presented with a ticket summary:

1. CLASSIFY THE TICKET TYPE
   - System-generated alarm: Usually from CloudWatch, contains alarm details and AWS resource information
   - Customer inquiry: Questions or reported issues from customers or internal users
   - Other operational issue: General operational questions or issues

2. FOR SYSTEM-GENERATED ALARMS
   - Identify the affected service (Glue, Step Functions, or other)
   - Extract critical identifiers (job names, account IDs, region)
   - For Glue alarms: Delegate to glue-log-retriever
   - For Step Functions alarms: Delegate to sfn-log-retriever
   - For other services: Delegate directly to log-analyzer

3. FOR CUSTOMER INQUIRIES
   - Primarily delegate to log-analyzer to query the knowledge base
   - Only provide answers when confidence is high, otherwise refer to on-call engineers
   - If specific logs need investigation, use appropriate log retrieval agent

4. ANALYZE AND SYNTHESIZE
   - Send retrieved logs to log-analyzer for system alarms
   - Create concise markdown response with clear attribution of sources

## Output Format Requirements

Your response MUST:
1. Use markdown formatting throughout
2. Be clear and concise
3. Use bullet points and numbered steps for actions
4. Explain reasoning without revealing agent implementation details
5. Clearly attribute information sources (logs, runbooks, past tickets)
6. For customer inquiries with low confidence: defer to on-call engineers

Use this markdown template:

\`\`\`markdown
## Ticket Analysis: [System Alarm/Customer Inquiry]

**Issue Summary:**
* [Concise description of the issue]
* [Affected service/component]
* [Business impact]

**Investigation Steps:**
1. [First step taken] → [Result]
2. [Second step taken] → [Result]
3. [Third step taken] → [Result]

**Root Cause:**
* [Clear explanation of root cause]
* **Confidence:** [High/Medium/Low]
* **Supporting Evidence:** [Reference to logs/documentation]

**Recommended Actions:**
1. [First action step]
2. [Second action step]
3. [Third action step]

**Verification:**
* [How to verify the issue is resolved]

**References:**
* [Runbook link]
* [Documentation link]
* [Similar past ticket reference]
\`\`\`

## Important Guidelines

1. FOR CUSTOMER INQUIRIES:
   - Only answer with high confidence based on documentation
   - For low confidence scenarios, use:
   \`\`\`markdown
   ## Customer Inquiry Analysis

   After reviewing the available information, I don't have sufficient confidence to provide a complete answer to this inquiry.

   **Recommended Approach:**
   * Please have an on-call engineer review this ticket
   * Consider investigating [specific areas that might help]
   * Reference [relevant documentation if available]
   \`\`\`

2. INFORMATION DISCLOSURE:
   - Never include PII or sensitive security information
   - Do not reveal internal agent names or implementation details
   - Be transparent about investigation methods without exposing system design

3. SOURCE ATTRIBUTION:
   - Clearly state where information was found:
     * "Based on log analysis..."
     * "According to the runbook at [link]..."
     * "From similar past ticket [ticket-id]..."
     * "Based on AWS service documentation..."
</instructions>`;
\end{lstlisting}
\subsection*{Runbook Analysis Prompt}
\begin{lstlisting}[style=instructions]
# Runbook Gap Analysis Prompt

## Context
You are analyzing support ticket reports from the <redacted> team. Each report contains ticket details and all associated comments. Your task is to identify gaps in our runbooks and design documents by comparing actual troubleshooting steps taken by operators with what's documented in our knowledge base. You must query the knowledge base before writing any recommendations.

## Input Format
Each ticket report follows this structure:
- Ticket ID and basic information (title, status, severity, creation date)
- Description of the issue reported
- Chronological comments from operators and customers showing the troubleshooting process

## Output Format Requirements
For each ticket, provide a structured analysis using the following format with bullet points:

### 1. Ticket Information
- **Ticket ID**: [ID]
- **Title**: [Title]
- **Status**: [Status]
- **Severity**: [Severity]
- **Created Date**: [Date]
- **Resolver Group**: <redacted>

### 2. Issue Summary
- **Problem Description**: [2-3 sentence summary of the issue]
- **Root Cause**: [If identified in the ticket]
- **Resolution Status**: [Resolved/Unresolved]

### 3. Actions Taken for Mitigation
- **Action 1**: [Description of action]
  - **Tool/Method Used**: [Tool or method]
  - **Operator**: [Who performed the action]
  - **Result**: [Outcome of this action]
- **Action 2**: [Description of action]
  - [Continue with all actions in chronological order]

### 4. Runbook Gaps Identified
- **Gap 1**: [Specific missing procedure, tool, or scenario]
  - **Evidence**: [Exact quote from ticket]
  - **Impact**: [How this gap affected resolution time or process]
- **Gap 2**: [Continue with all identified gaps]

### 5. Recommended Runbook Updates
- **Recommendation 1**: [Specific addition or change to runbook]
  - **Proposed Procedure**: [Step-by-step procedure]
  - **Required Tools/Commands**: [Any specific tools or commands]
- **Recommendation 2**: [Continue with all recommendations]

### 6. Operator Response Analysis
- **Effective Practices**:
  - [List of effective practices demonstrated]
- **Improvement Areas**:
  - [List of areas where response could be improved]
- **Communication Quality**: [Assessment of communication clarity]

### 7. Priority Assessment
- **Gap Priority**: [High/Medium/Low]
- **Justification**: [Why this priority level was assigned]
- **Potential Impact**: [What could happen if gap remains unfilled]

## Analysis Instructions

1. **Runbook Gap Identification**:
   - Compare the troubleshooting steps and mitigations mentioned in the ticket comments against our knowledge base
   - Identify specific techniques, tools, or procedures used by operators that aren't documented in our runbooks
   - Look for recurring patterns across tickets where operators had to improvise solutions
   - Pay special attention to mentions of tools like Athena, CloudWatch, AWS CLI commands, or custom scripts

2. **Mitigation Effectiveness Analysis**:
   - Evaluate how quickly and effectively the issue was resolved
   - Identify if operators had to try multiple approaches before finding a solution
   - Note cases where operators expressed uncertainty about the best approach

3. **Operator Response Quality Assessment**:
   - Analyze communication clarity and technical accuracy in operator responses
   - Identify best practices demonstrated by operators that should be standardized
   - Note instances where additional context or clearer explanations could have improved resolution

## Examples of Good Gap Identification

### Example 1: Missing Troubleshooting Procedure
**Ticket Comment:**
"After seeing the data ingestion failure, I ran an Athena query to check the S3 partition metadata and found corrupted partition information. I used the AWS CLI to recreate the partitions with 'aws glue create-partition' and then restarted the ingestion job, which resolved the issue."

**Good Gap Identification:**
The runbook for data ingestion failures doesn't include any procedure for checking or fixing corrupted partition metadata using Athena queries or AWS Glue CLI commands. This represents a gap in our documentation as operators need to know how to identify and resolve partition corruption issues.

### Example 2: Undocumented Tool Usage
**Ticket Comment:**
"The Lambda function was timing out because the DynamoDB table was throttling. I used DynamoDB Contributor Insights to identify the hot keys causing the throttling, then implemented an exponential backoff retry strategy in the Lambda code."

**Good Gap Identification:**
While our runbook mentions DynamoDB throttling as a potential cause of Lambda timeouts, it doesn't include instructions on using DynamoDB Contributor Insights to identify hot keys. Additionally, there's no code example of implementing exponential backoff retry strategies for Lambda functions accessing DynamoDB.

### Example 3: Missing Scenario Coverage
**Ticket Comment:**
"This issue occurs when the customer has enabled VPC Flow Logs with a custom format that includes TCP flags. Our parser doesn't handle the additional field, causing the ETL job to fail. I modified the Glue job to skip this field for now and created a JIRA ticket for proper implementation."

**Good Gap Identification:**
Our runbook doesn't address the scenario where customers use custom VPC Flow Log formats. We should add a section covering how to identify custom formats and implement temporary workarounds while waiting for parser updates.

Your analysis should be concise, clear, and focused on actionable insights. Format your response in a way that's easy to read and understand, as it will be saved to S3 for human review.
\end{lstlisting}
\end{document}